\documentclass[runningheads]{llncs}
\usepackage{graphicx}

\usepackage{tikz}
\usepackage{comment}
\usepackage{amsmath,amssymb} 
\usepackage{color}

\usepackage[accsupp]{axessibility}  

\usepackage{algpseudocode}
\usepackage{algorithm}
\usepackage{bm}
\usepackage{subcaption}

\usepackage{booktabs}
\usepackage{threeparttable}

\begin{document}
\pagestyle{headings}
\mainmatter
\def\ECCVSubNumber{7566}  

\title{ARDIR: Improving Robustness using Knowledge Distillation of Internal Representation} 

\titlerunning{ARDIR}
%
\author{Tomokatsu Takahashi\inst{1}\and
Masanori Yamada\inst{1} \and
Yuuki Yamanaka\inst{1} \and
Tomoya Yamashita\inst{1}}
\authorrunning{T. Takahashi et al.}
%
\institute{NTT Social Informatics Laboratories}
\maketitle

\begin{abstract}
Adversarial training is the most promising method for learning robust models against adversarial examples. 
A recent study has shown that knowledge distillation between the same architectures is effective in improving the performance of adversarial training.
Exploiting knowledge distillation is a new approach to improve adversarial training and has attracted much attention.
However, its performance is still insufficient.
Therefore, we propose Adversarial Robust Distillation with Internal Representation~(ARDIR) to utilize knowledge distillation even more effectively.
In addition to the output of the teacher model, ARDIR uses the internal representation of the teacher model as a label for adversarial training.
This enables the student model to be trained with richer, more informative labels.
As a result, ARDIR can learn more robust student models.
We show that ARDIR outperforms previous methods in our experiments.
\end{abstract}

\section{Introduction}
\label{sec1}
Deep neural networks (DNNs) have made remarkable achievements in a wide range of areas such as image processing~\cite{krizhevsky2012imagenet}, speech recognition~\cite{hinton2012deep} and cyber security~\cite{xu2018unsupervised,david2015deepsign}.
However, DNNs are vulnerable to adversarial examples which are data with perturbations that are imperceptible to humans, and can be easily fooled, leading to misclassification~\cite{szegedy2013intriguing,goodfellow2014explaining}.
This is problematic because vulnerability to adversarial examples is an obstacle to adapting DNNs to reliability-critical fields such as automated driving.
Hence, various defensive methods have been proposed to date.

Currently, the most promising defensive method against adversarial examples is adversarial training, which uses adversarial examples as training data to learn robust models~\cite{kurakin2016adversarial,madry2017towards}.
The most famous Adversarial Training method~\cite{madry2017towards} creates a robust model by learning the adversarial example generated by Projected Gradient Descent~(PGD) as training data.
This method is known as Standard Adversarial Training~(SAT), and various improvements have been proposed on the basis of it~\cite{zhang2019theoretically,wang2019convergence,rice2020overfitting,wu2020adversarial}.
However, despite the various repeated improvements, DNNs are still vulnerable to adversarial examples in order to adapt to realistic problems.
Therefore, improvements through new approaches are necessary.

Recent studies have shown that adapting knowledge distillation~\cite{hinton2015distilling} to adversarial training can lead to improved robustness of the models~\cite{goldblum2020adversarially,chen2021robust,zhu2022reliable}.
Knowledge distillation is a method of training a small student model using the output of a large and well-performing teacher model as a label~\cite{hinton2015distilling}.
Similarly, knowledge distillation is used in adversarial training to transfer the robustness of a large teacher model to a small student model~\cite{goldblum2020adversarially,zi2021revisiting}.
Adversarial Robust Distillation~(ARD)~\cite{goldblum2020adversarially} is the first method to apply distillation techniques to adversarial training.
Although their main goal is to reduce the size of the robust model, their paper contains a notable experimental result that the student model has higher robust accuracy (accuracy for adversarial examples) than the teacher model when the teacher and student models have the same sized architecture.
This result implies that knowledge distillation effectively improves the robustness of adversarial trained models.
On the basis of this results, several methods have been proposed to improve the robustness of adversarial trained models~\cite{chen2021robust,zhu2022reliable}.
However, the robust accuracy against adversarial examples of these methods is still insufficient in order to adapt DNNs to reliability-critical fields.

To further improve the robust accuracy of student models, we focused on knowledge distillation using the internal representation of the teacher model.
Since knowledge distillation utilizes an informative label that is the output of the teacher model, it can improve the accuracy of the student model.
Thus, we consider utilizing more informative labels which are internal representations of the teacher model.
In fact, in clean training, the use of internal representations for knowledge distillation has been successful in improving accuracy~\cite{romero2014fitnets}.

Furthermore, we consider the best combination of teacher model and teacher data to utilize the internal representation of the teacher model more effectively.
In the adversarial training model, overfitting occurs against the learned adversarial example, and test robust accuracy decreases as the training progresses beyond a certain point~\cite{rice2020overfitting}.
As a result, the robust generalization gap, the difference between training robust accuracy and test robust accuracy, becomes larger as learning progresses.
On the other hand, since test clean accuracy (accuracy against test clean data) increases as training progresses, the clean generalization gap, the difference between training clean accuracy and test clean accuracy, is smaller than that for the adversarial example.
On the basis of these findings, we believe that the internal representation of the teacher model for clean data is more generic than those for the adversarial example.
Therefore, we thought that the internal representation when clean data input the robust teacher model improves the robust accuracy of the student model. 

\subsection{Contributions}
\begin{figure}[t]
    \centering
    \includegraphics[height=5cm]{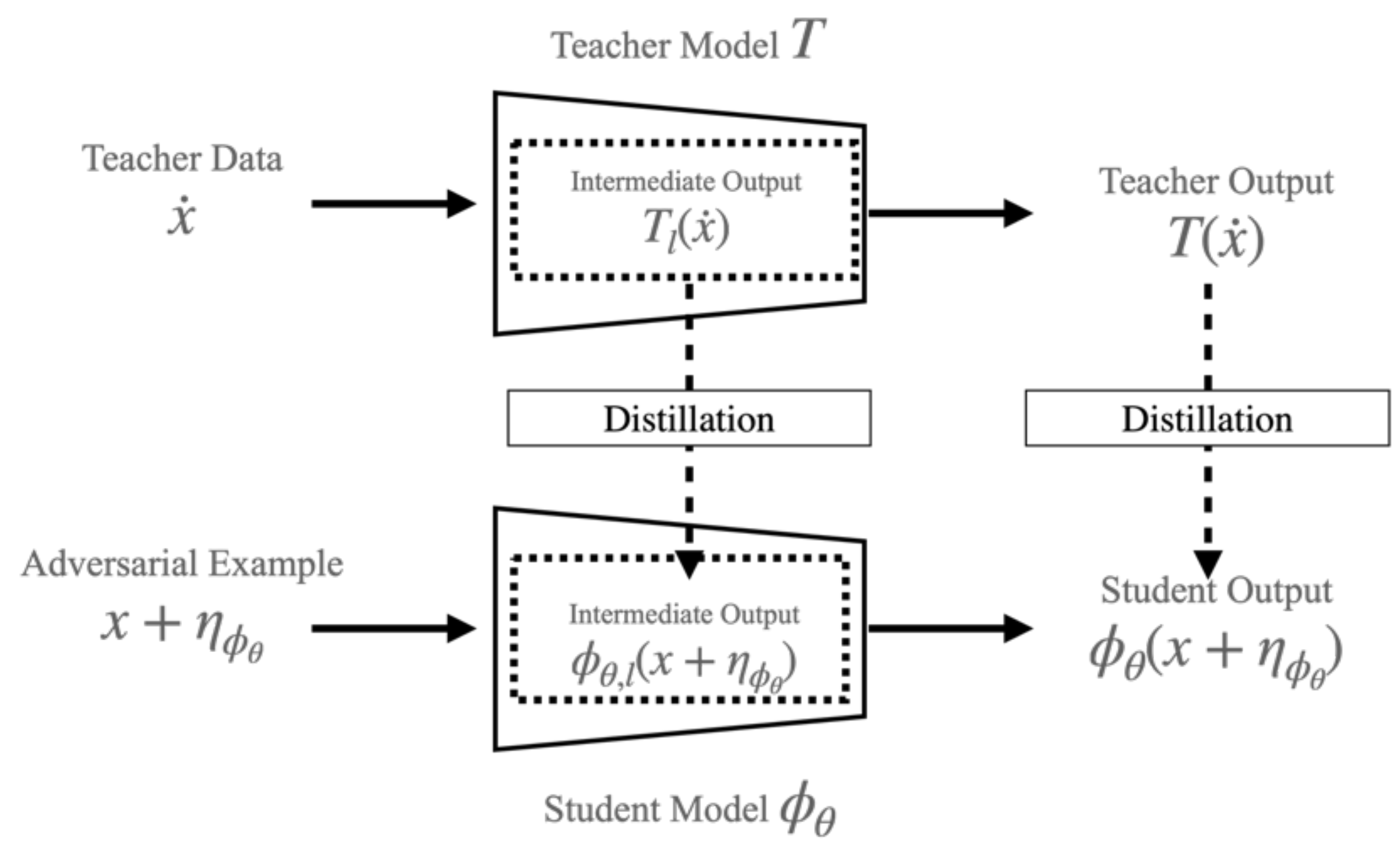}
    \caption{An overview of ARDIR. ARDIR trains the student model to bring the output of the student model and the teacher model closer, and the intermediate output~(i.e. internal representation) of the student model and the teacher model closer, respectively.}
    \label{overview}
\end{figure}

Motivated by the above reasons, we propose Adversarial Robust Distillation with Internal Representation~(ARDIR).
An overview of ARDIR is shown in Figure~\ref{overview}.
Our goal is to obtain a robust student model against adversarial examples.
To this end, ARDIR adapts knowledge distillation more effectively to adversarial training by using an internal representation in addition to the output of the teacher model.
As mentioned above, we believe that the performance of the model can be further improved by learning the internal representation of clean data.
On the basis of this idea, we measure and learn the difference between the internal representations of the teacher and student models using Learned Perceptual Image Patch Similarity~(LPIPS)~\cite{zhang2018unreasonable}.
LPIPS is a measure of similarity between images using the internal representation of the model and is known to be close to human perception.
Clean data and adversarial examples are very close in human perception.
Thus, we can train a robust model that recognizes adversarial examples as similar to clean data by minimizing the LPIPS between clean data input to the teacher model and adversarial examples input to the student model.

Furthermore, we investigate the best combination of the teacher model and the teacher data for adversarial training using knowledge distillation.
The existing method ARD uses the robust model as the teacher model and clean data as the teacher data.
However, it is not clear whether this combination is the most effective or not.
For example, since the clean model which is trained with clean data has high clean accuracy, good output and internal representation can possibly be obtained when clean data is used as input.
In addition, for the robust model, it may be possible to obtain a label that effectively transfers the robustness of the teacher model to the student model by using an adversarial example as the teacher data.
Therefore, we investigate not only the combination of the robust model and clean data but also the case where the clean model and adversarial examples are used as teachers.

Finally, we conduct the experiments on ARDIR with multiple datasets and attacks to evaluate its effectiveness.
We show that ARDIR can learn student models that are more robust than previous methods.
As expected, ARDIR achieved its best performance when the internal representations of clean data input to the robust model were used for training as labels.

Our contributions in this paper are summarized as follows:
\begin{itemize}
\item \textbf{Proposed ARDIR:}
We propose a novel adversarial training using knowledge distillation of internal representation, called Adversarial Robust Distillation with Internal Representation~(ARDIR).
ARDIR uses the high quality internal representation of clean data input to the teacher model.
This enables us to learn a student model with higher robust accuracy than the teacher model.

\item \textbf{Investigate the combination of teacher model and teacher data:} 
We perform evaluation experiments on combinations of teacher models and data.
As a result, we show that the combination of the robust model and clean data performs the best in the proposed method ARDIR.

\item \textbf{Confirm that ARDIR achieves higher robustness than SOTA:}
We show that the ARDIR outperforms previous methods including Introspective Adversarial Distillation (IAD)~\cite{zhu2022reliable} which is the state of the art.
To this end, we conduct experiments on our proposed method with multiple datasets and attacks in Sec.~\ref{sec_experiments}. 
\end{itemize}

To the best of our knowledge, our proposed method ARDIR is the first to adopt the knowledge distillation of internal representation to adversarial training.
ARDIR, distillation with internal representations, can train more robust models than previous methods.

\section{Preliminary}
\label{sec_preliminary}
In this section, we formally define the notations and explain ARD~\cite{goldblum2020adversarially} and related knowledge.
First, we describe the adversarial example and its typical attack methods in Sec.~\ref{sec2.1}.
We then outline adversarial training in Sec.~\ref{sec2.2}.
Finally, the previous research ARD is explained in Sec.~\ref{sec2.3}.

\subsection{Adversarial Example}
\label{sec2.1}
Adversarial examples~\cite{szegedy2013intriguing} are malicious data designed to cause misclassification in DNN and are created by applying imperceptible perturbations to the input data.
We describe two types of typical adversarial examples.

Fast Gradient Sign Method (FGSM)~\cite{goodfellow2014explaining} is a most basic attack method that uses the gradient of the model's loss function to generate an adversarial example that increases the loss function.
Given the clean data $\bm{x}\in \mathbb R^d$, the label $\bm{y} \in \mathbb R^k$, and target model $\phi_{\bm{\theta}}$ (parameterized by $\bm{\theta}$) and the loss function $\ell$, FGSM generate an adversarial example as follows:
\begin{align}
  \label{fgsm}
  \bm{x}+\bm{\eta}_{\phi_{\bm{\theta}}} &= \bm{x} + \epsilon~ {\rm sign} \left( \nabla_{\bm{x}} \ell  \left(\phi_{\bm{\theta}} \left(\bm{x} \right) ,\bm{y}\right) \right) ,
\end{align}
where $\bm{\eta}_{\phi_{\bm{\theta}}}$ 
is adversarial perturbation that depends on the model $\phi_{\bm{\theta}}$, $\bm{x}+\bm{\eta}_{\phi_{\bm{\theta}}}$ 
is adversarial example, ${\rm sign}$ is function that extracts the sign of vector, and $\epsilon$ 
is the magnitude of the perturbation to be added to $\bm{x}$.

Projected Gradient Descent (PGD)~\cite{madry2017towards} is a powerful attack method derived from FGSM.
In FGSM, the adversarial example is created by adding a perturbation of $\epsilon$ magnitude to $\bm{x}$ at a time.
On other hand, PGD iteratively updates the adversarial example for each step and added perturbation of $a$, where $a$ is stepsize.
At each step, if the magnitude of the perturbation exceeds $\epsilon$, it is clipped into set $B$ by the projection function $\prod_B$, where $B$ is the adversarial set is defined as $B = \{x + \bm{\eta}_{\phi_{\bm{\theta}}} | \|\bm{\eta}_{\phi_{\bm{\theta}}} \|_p \leq \epsilon\}$. 
PGD generate an adversarial example as follows:
\begin{align}
  \bm{x} + \bm{\eta}^{i+1}_{\phi_{\bm{\theta}}}  &=  \prod_B \left(\bm{x} + \bm{\eta}^{i}_{\phi_{\bm{\theta}}} + a~{\rm sign}\left(\nabla_{\bm{x}} \ell\left(\phi_{\bm{\theta}}\left(\bm{x}+\bm{\eta}^{i}_{\phi_{\bm{\theta}}}\right),\bm{y}\right)\right)\right),
  \label{pgd}
\end{align}
where $\bm{\eta}^{0}$ is the initial perturbation, which is generally random noise.

\subsection{Adversarial Training}
\label{sec2.2}
In this section, we show the definition of adversarial training~\cite{kurakin2016adversarial}.
Adversarial training is a training method to create a robust model by incorporating adversarial examples as training data. 
Adversarial training optimizes the parameters $\bm{\theta}$ as follows
\begin{align}
  \min_{\bm{\theta}} \mathbb E_{(\bm{x},\bm{y}) \sim \bm{D}} \left[\max_{ \bm{\eta}_{\phi_{\bm{\theta}}} \in B} \ell\left(\phi_{\bm{\theta}}\left(\bm{x} + \bm{\eta}_{\phi_{\bm{\theta}}}\right),\bm{y}\right)\right] 
  \label{AT},
\end{align}
where  $D$ is the training dataset.
In most cases, the adversarial example $\bm{x}+ \bm{\eta}_{\phi_{\bm{\theta}}}$ used in adversarial training is created using PGD.

\subsection{Adversarial Robust Distillation~(ARD)}
\label{sec2.3}
ARD~\cite{goldblum2020adversarially} is the first method to apply the concept of knowledge distillation to adversarial training.
First, we introduce normal knowledge distillation~\cite{hinton2015distilling}.
In many cases, knowledge distillation is used to compress the size of a model, using the output of a large teacher model as labels to train a small teacher model. 
Given the trained teacher model $T$, knowledge distillation optimizes the parameters of the student model $\bm{\theta}$ for the following equation~\ref{distilluation}
\begin{align}
  \min_{\bm{\theta}} \mathbb E_{(\bm{x},\bm{y}) \sim D} \left[\alpha t^2 {\rm KL}\left(\sigma_{t}\left(\phi_{\bm{\theta}}\left(\bm{x} \right)\right),\sigma_{t}\left(T\left(\bm{x}\right)\right)\right)
    + \left(1-\alpha\right)\ell\left(\phi_{\bm{\theta}}\left(\bm{x} \right),\bm{y}\right)\right] ,
  \label{distilluation}
\end{align}
where $\sigma_{t}$ is the softmax with temperature $\sigma_{t}(\bm{z})=\frac{\exp(\bm{z_i}/t)}{\sum_{i=1}\exp(\bm{z_i}/t)}$ , $t$ is the temperature constant, KL is the Kullback-Leibler divergence, and $T\left(\bm{x}\right)$ is the logits of $T$ when $\bm{x}$ is input.
The first term of Equation~\ref{distilluation} is performs learning to match the output of the student model to the output of the teacher model, and the second term performs learning with a true label, as in clean training.
$\alpha=[0,1]$ is the hyperparameter, which determines the ratio of the two terms.

ARD is a simple extension of normal knowledge distillation to adversarial training that uses the robust model learned in adversarial training as the teacher model.
ARD inputs clean data to the robust teacher model and obtains its logits.
Then, it performs adversarial training on the student model using the logits as labels.
In this way, the student model can be trained to outperform the teacher model.
ARD optimizes the parameters of student model $\bm{\theta}$ for the following equation~\ref{ARD}
\begin{align}
  \min_{\bm{\theta}} \mathbb E_{(\bm{x},\bm{y}) \sim D} \left[\alpha t^2 {\rm KL}\left(\sigma_{t}\left(\phi_{\bm{\theta}}\left(\bm{x} + \bm{\eta}_{\phi_{\bm{\theta}}}\right)\right),\sigma_{t}\left(T\left(\bm{x}\right)\right)\right) 
  \right. \nonumber \\ 
    \left.
    + \left(1-\alpha\right)\ell\left(\phi_{\bm{\theta}}\left(\bm{x} + \bm{\eta}_{\phi_{\bm{\theta}}}\right),\bm{y}\right)\right], 
  \label{ARD}
\end{align}
where $T$ represents a trained teacher model as in general distillation, but in the case of ARD, it is a robust model because ARD aims to inherit the robustness of the teacher model.
The first term of Equation~\ref{ARD} is for learning to match the output of the student model to the output of the teacher model, and the second term is for learning with a true label, as in normal learning.
On the basis of the settings in the ARD paper, the hyperparameter $\alpha$, which determines the ratio of the two terms, is always $\alpha=1$ in this paper.

\section{Proposed Method}
\label{sec_proposed_method}
In this section, first, we present adversarial training using knowledge distillation of internal representation, called ARDIR in Sec.~\ref{sec_ardir}.
Next, we discuss the combinations of teacher models and their input data in Sec.~\ref{sec_teachar_combination}.

\subsection{ARDIR}
\label{sec_ardir}
ARD~\cite{goldblum2020adversarially} adapts a simple knowledge distillation method to adversarial training, using only the output of the teacher model. 
On the other hand, we believe that knowledge distillation using internal representations in addition to the output can further improve adversarial robustness.
Therefore, we propose adversarial training using knowledge distillation of internal representation.

We think the use of internal representations will be effective for two reasons.
The first reason is the increase in the amount of information as labels.
The one-hot label used in the usual supervised learning setting only has information about the correct class.  
On the other hand, the output of the teacher model has more detailed information about the relationship between each class and the input data~\cite{hinton2015distilling}.
This is one reason why knowledge distillation can improve performance.
We can further use more informative labels by using the internal representation.
The second reason is the availability of high quality features for clean data in robust models.
Adversarial training tends to overfit the adversarial examples~\cite{rice2020overfitting}.
As a result, after a certain number of training
loops, test robust accuracy for the adversarial example decreases.
On the other hand, the test clean accuracy for clean data improves as the training loop progresses. 
As a result, the robust model has a small clean generalization gap and a large robust generalization gap for adversarial examples shown in Table~\ref{gap}.
The experimental settings are described in Sec.~\ref{sec_setup}.
\begin{table}[t]
\begin{center}
\caption{The generalization performance of the robust model.}
\begin{tabular}{@{}lrrrr@{}}
\toprule
& \multicolumn{2}{c}{Cifar10} & \multicolumn{2}{c}{SVHN} \\ \midrule
 & \multicolumn{1}{c}{Clean} & \multicolumn{1}{c}{PGD} & \multicolumn{1}{c}{Clean} & \multicolumn{1}{c}{PGD} \\ \midrule
Train Acc          & 0.8528       & 0.5927       & 0.9376      & 0.6544     \\
Test Acc           & 0.8094       & 0.5229       & 0.9067      & 0.5247     \\
Gap~(Train - Test) & 0.0434       & 0.0698       & 0.0309      & 0.1297     \\
Ratio~(Test/Train) & 0.9491       & 0.8822       & 0.9670      & 0.8018     \\ \bottomrule
\end{tabular}
 \label{gap}
\end{center}
\end{table}

This result indicates the cause of the difference between the robust and clean generalization gap is the features learned by the robust model for clean data are intrinsic, unaffected by trivial differences between data.
Thus, we believe that ARDIR can improve the robust accuracy by using the feature of clean data learned by the robust model.

In addition, since clean data and adversarial examples have the same features in human perception, the features handled by the model should also be the same.
Therefore, we thought that distilling the internal representation for clean data in the teacher model would improve the performance.
ARDIR optimizes the following loss function
\begin{align}
  \min_{\bm{\theta}} \mathbb E_{\left(\bm{x},\bm{y}\right) \sim D} \left[\left(1-\beta\right) {\rm KL}\left(\sigma_{t}\left(\phi_{\bm{\theta}}\left(\bm{x} + \bm{\eta}\right)\right),\sigma_{t}\left(T\left(\bm{\dot{x}}\right)\right)\right) 
  \right. \nonumber \\
  \left.
  + \beta R\left(\phi_{\bm{\theta}}\left(\bm{x} + \bm{\eta}\right),T\left(\bm{\dot{x}}\right)\right)\right],
  \label{ARDIR}
\end{align}
where $R$ represents the distance function between the intermediate outputs of $\phi_{\bm{\theta}}(\bm{x} + \bm{\eta})$ and $T(\bm{\dot{x}})$.
$T$ is the teacher model, which is either the clean model or the robust model.
$\bm{\dot{x}}$ is teacher data, and either clean data $\bm{x}$ or adversarial example $\bm{x}+\bm{\eta}_{T}$ is chosen.

In this paper, we use LPIPS~\cite{zhang2018unreasonable} as a distance function to measure the distance of intermediate outputs. 
LPIPS is a measure of the distance between images using the intermediate output of the model, and it is known that LPIPS is close to human perception when the model has been properly trained.
In ARDIR, LPIPS is used to measure the difference between the internal representations of the teacher and student models.
There are two reasons why the LPIPS is suitable as a distance between internal representations in our proposed method
The first reason is that LPIPS enables all layers to be compared equally regardless of the size of the layer since it is calculated for each intermediate output vector normalized by the width of the layer and the size of the output.
The second reason is that LPIPS is closer to human perception.
The adversarial example and clean data are very similar in terms of human perception, but the model discriminates them as different images.
Therefore, we learn to match the model with human perception by learning to bring the LPIPS closer to the clean data of the teacher model.
LPIPS is defined as
\begin{align}
  \label{lpips}
  {\rm LPIPS}\left(\bm{x_1},\bm{x_2},\phi \right) &= \| \xi\left(\bm{x_1},\phi\right) - \xi\left(\bm{x_2},\phi\right) \|_2, \\
  \xi\left(\bm{x}\right) &= \left(\frac{\hat \phi_l\left(\bm{x}\right)}{\sqrt{W_1 H_1}},\dots,\frac{\hat \phi_L \left(\bm{x}\right)}{\sqrt{W_L H
      _L}} \right), \\
  \hat \phi_{l} \left(\bm{x}\right) &= \left( \hat \phi_{l1} \left(\bm{x}\right) ,\dots,\hat \phi_{lC} \left(\bm{x}\right)\right),  \\
  \hat \phi_{lc} \left(\bm{x}\right) &= \frac{\phi_{lc} \left(\bm{x}\right)}{\sum^{W}_{w}\sum^{H}_{h} \phi_{lcwh} \left(\bm{x}\right)} ,
\end{align}
where $\phi_{lc}(x)$ is the intermediate output vector of channel $c$ in layer $l$, 
and $\hat \phi_{lc}(x)$ is a vector of $\phi_{lc}$ normalized for each channel $c$.
Also, $\hat \phi_l(x)$ is a vector of $\phi_{lc}(x)$ for channel $c$, where $\xi(x)$ is a vector of $\hat \phi_l(x)$ normalized by the layer size and further ordered by the number of layers.
As mentioned above, LPIPS normalizes the intermediate output vector of each layer by the layer size and output size.
Then, it calculates the L2 norm for the normalized vector.
In the proposed method, it is used to calculate the distance between the intermediate outputs of the teacher model and the student model.
Thus, the LPIPS is calculated as
\begin{eqnarray}
  {\rm LPIPS_{ARDIR}}\left(\bm{x},\bm{\dot{x}},\phi_{\bm{\theta}},T\right) = \| \xi\left(\bm{x},\phi_{\bm{\theta}}\right) - \xi\left(\bm{\dot{x}},T\right) \|_2 .
  \label{lpips_ours}
\end{eqnarray}
Finally, the Loss function of ARDIR incorporating LPIPS defined as, 
\begin{align}
  \min_{\bm{\theta}} \mathbb E_{\left(\bm{x},\bm{y}\right) \sim D} \left[\left(1-\beta\right) {\rm KL}\left(\sigma_{t}\left(\phi_{\bm{\theta}}\left(\bm{x} + \bm{\eta}\right)\right),\sigma_{t}\left(T\left(\bm{\dot{x}}\right)\right)\right) 
  \right. \nonumber \\
  \left.
  + \beta {\rm LPIPS_{ARDIR}}\left(\bm{x},\bm{\dot{x}},\phi_{\bm{\theta}},T\right)\right] .
  \label{ARDIR_LPIPS}
\end{align}

\subsection{Teacher Combinations}
\label{sec_teachar_combination}
In this section, we consider the combination of a teacher model and teacher data in adversarial training using knowledge distillation.
The previous research ARD uses Robust Model trained by Adversarial Training as the teacher model and Clean Data as the teacher data.
Although the ARD focuses only on the combination of robust model and clean data, there are actually four possible combinations.
Each combination pattern is explained below.

{\bf Clean Model + Clean Data (CC):}
In the setting of CC, the output or internal representation of the clean model with clean data input is used as a label.
This is the same setting as the normal distillation~\cite{hinton2015distilling} in clean training.
The clean model classify more accurate than the robust model for clean data, thus it may produce better output and internal representation than the robust model when clean data is input.
On the other hand, since the clean model is vulnerable to the adversarial example, we do not expect much robustness inheritance from the teacher model to the student model.

{\bf Robust Model + Clean Data (RC):}
RC is the same setting as in the previous research  ARD~\cite{goldblum2020adversarially}.
It is also the setting that is expected to produce the best performance in the proposed method.
Robustness is expected to be inherited from robust model to student model.
Furthermore, as mentioned in Sec.~\ref{sec_ardir}, ARDIR can use a good quality internal representation of the clean data input.

{\bf Robust Model + Adversarial Example (RA):}
There is no method that uses RA alone, but it is sometimes used in combination with other combinations in some previous studies. ~\cite{chen2021robust,zhu2022reliable}
Since the robust model is trained to classify the adversarial example, it is expected to produce labels that better express the robustness of the robust model.

{\bf Clean Model + Adversarial Example (CA):}
CA inputs the adversarial example to the clean model.
However, the accuracy of the clean model for the adversarial example is almost zero.
We do not consider this combination since it does not yield a valid output.
We examined each of the teacher model and data combinations by conducting experiments in Sec.~\ref{sec_experiments}.

\subsection{Algorithm}
Algorithm~\ref{ARDIR_algorithm} shows the algorithm of the proposed method ARDIR.
Firstly, ARDIR determines the teacher data to be input into the teacher model.
For teacher combinations CC and RC, the teacher data $\bm{\dot{x}}=\bm{x}$, and for RA, $\bm{\dot{x}}=\bm{x}+\bm{\eta}_{T}$.
Where $\bm{x}+\bm{\eta}_{T}$ is the Adversarial Example made for the teacher model $T$.
Next, ARDIR obtains Teacher Output $\bm{\dot{y}}$ as a label for learning the student model.
At this time, if the class indicated by $\bm{\dot{y}}$ is wrong, the $\bm{\dot{y}}$ is replaced by the one-hot label $\bm{y}$.
Then, the loss function $\ell_{ARDIR}$ is calculated based on the expression~\ref{ARDIR}, and the gradient is used to update the parameter $\bm{\theta}$ of the student model.
\begin{algorithm}[t]
    \caption{ARDIR}
    \label{ARDIR_algorithm}
    \begin{algorithmic}
    \Require Training Dataset $D = \{(\bm{x}_i,\bm{y}_i)\}^n_{i=1}$, Student Model $\phi_{\theta}$,Teacher Model $T$, learning rate $\gamma$, Number of epoch $N$, batch size $m$, Number of Batches $M$, adjustable parameter $\beta$, temperature constant $t$
    \Ensure Robust Student Model $\phi_{\theta^{*}}$
    \For{${\rm Epoch}=1,\dots,N$}
    \For{${\rm Batch}=1,\dots,M$}
    \State Compute Adversarial Perturbation $\bm{\eta}_{\phi_\theta}$ for $\bm{x}_i \in {\rm Batch}$ using PGD in Eq.\ref{pgd}
    \If{ARDIR~(CC) or ARDIR~(RC)}
    \State Teachar data $\bm{\dot{x}} \leftarrow \bm{x}$
    \State Teachar output $\bm{\dot{y}} \leftarrow T(\bm{x})$
    \ElsIf{ARDIR~(RA)}
    \State Compute Adversarial Perturbation $\bm{\eta}_{T}$ for $\bm{x}_i \in {\rm Batch}$ using PGD in Eq.\ref{pgd}
    \State Teachar data $\bm{x}^T \leftarrow \bm{x}+\bm{\eta}_{T}$
    \State Teachar output $\bm{\dot{y}} \leftarrow T(\bm{x}+\bm{\eta}_{T})$
    \EndIf
    \For{i=1,\dots,m}
    \If{$\max(\bm{\dot{y}}_i) \neq \bm{y_i}$}
    \State $\bm{\dot{y}}_i \leftarrow \bm{y_i}$
    \EndIf
    \EndFor
    \State $\ell_{ARDIR}(\bm{x}\!+\!\bm{\eta},\theta) \!=\! \left(1-\!\beta\right) {\rm KL}\left(\sigma_{t}\left(\phi_{\theta}\left(\bm{x} \!+\! \bm{\eta}\right)\right),\sigma_{t}\left(T\left(\bm{\dot{x}}\right)\right)\right) \!+\! \beta LPIPS_{ARDIR}\left(\bm{x},\bm{\dot{x}},\phi_{\theta},T\right)$
    \State $\theta \leftarrow \theta - \gamma\nabla_{\theta}\ell_{ARDIR}(\bm{x}+\bm{\eta},\theta)$
    \EndFor
    \EndFor
    \end{algorithmic}
\end{algorithm}

\section{Experiments}
\label{sec_experiments}
\subsection{Setup}
\label{sec_setup}
In this section, we describe the experimental setup for evaluating our proposed method ARDIR.
In this experiment, we use CIFAR10~\cite{krizhevsky2009learning} and SVHN~\cite{netzer2011reading}.
We also use PreActResNet-18~\cite{he2016identity} as a teacher model and a student model.

{\bf Attack methods:}
To evaluate the robustness of each defense methods, we use FGSM~\cite{goodfellow2014explaining}, PGD~\cite{madry2017towards} and AutoAttack~\cite{croce2020reliable} as attack methods.
AutoAttack selects the most effective adversarial example against a model among multiple attacks and is used as a benchmark for robustness.
We set the magnitude of perturbation $\epsilon=8/255$, the PGD step size $a=2/255$ (CIFAR-10) or $a=1/255$ (SVHN), PGD iteration numbers $k=10$ (Training) or $k=20$ (Test).

{\bf Defense methods:}
To evaluate our proposed method, we compare Standard Adversarial Training~(SAT)~\cite{madry2017towards}, Adversarial Robust Distillation~(ARD)~\cite{goldblum2020adversarially} and Introspective Adversarial Distillation (IAD)~\cite{zhu2022reliable} with the proposed method.
SAT and ARD are as described in sec~\ref{sec_preliminary}.
IAD is an extension of ARD and it archives the state-of-the-art robust accuracy.
Within the IAD paper, IAD-1 and IAD-2 are proposed as variations of IAD.
We note that the hyperparameter settings are not published for IAD-2, so we cite the reference values in their paper.

All defense methods are trained for 200 epochs using Stochastic Gradient Decent~(SGD) with momentum 0.9, weight decay $5 \times 10^{-4}$, and an initial learning rate of 0.1 that is divided by 10 at the 100th and 150th epoch.
Then, the model at the epoch that shows the best test robust accuracy against PGD is used as the final result.
In ARD and ARDIR, we set temperature constant $t=1$ or $t=30$.
The teacher model is selected from the clean model and the robust model which is trained by SAT.
The test robust accuracy~(PGD) of the robust teacher models are 52.29$\%$~(CIFAR10) and 52.47$\%$~(SVHN).
Therefore, if the output obtained from the teacher model is not correct, we replace the label with the correct one-hot label and train it.
For teacher data, an adversarial example or clean data is used.
The experiment also examines which is best for CC, RC, or RA.
The adversarial example used for training each method is generated by using PGD.

\begin{figure}[t]
   \begin{minipage}[b]{0.32\linewidth}
    \centering
    \includegraphics[height=3.3cm]{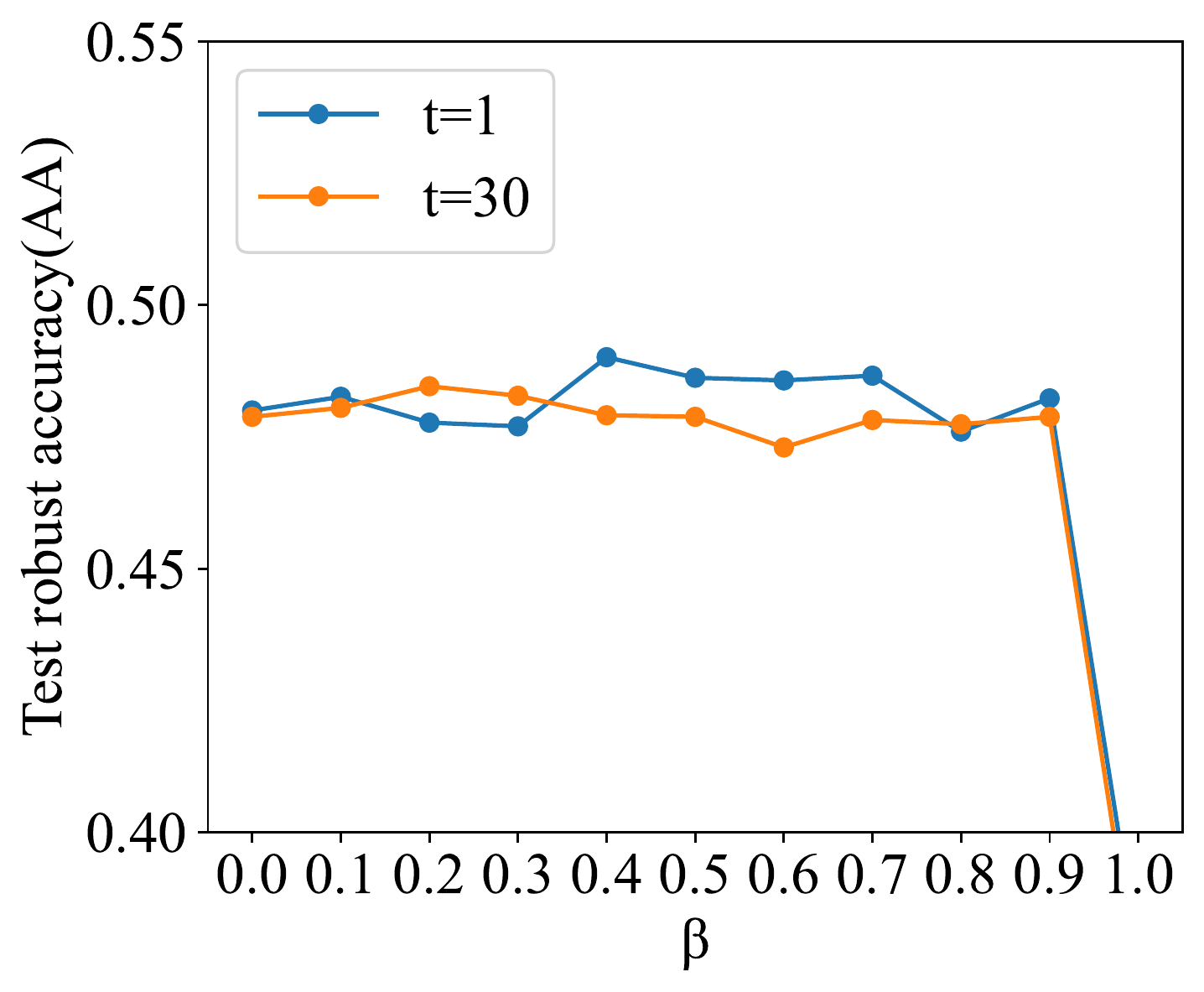}
    \subcaption{CC}
    \label{cifar10_CC}
  \end{minipage}
   \begin{minipage}[b]{0.32\linewidth}
    \centering
    \includegraphics[height=3.3cm]{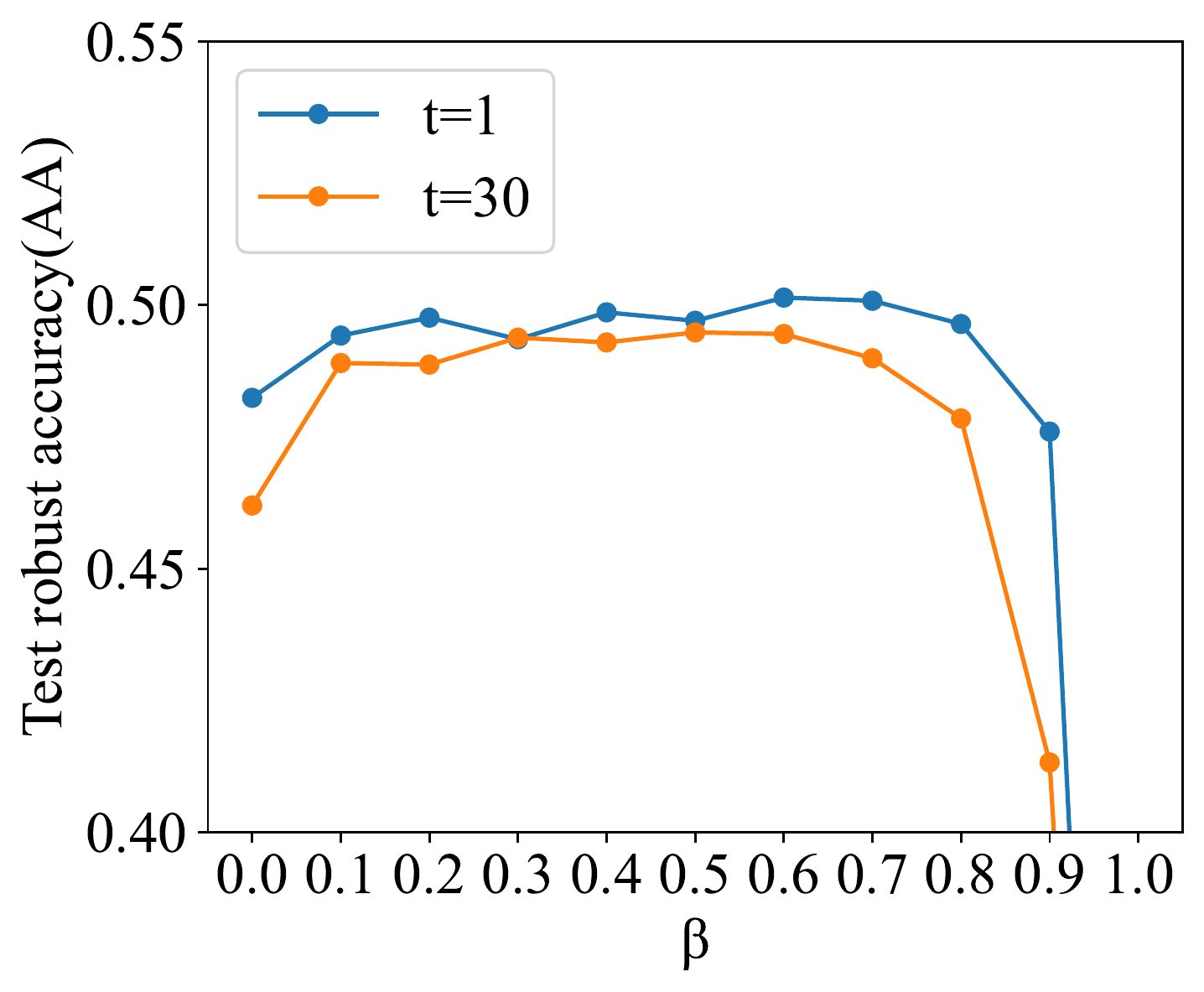}
    \subcaption{RC}
    \label{cifar10_RC}
  \end{minipage}
  \begin{minipage}[b]{0.32\linewidth}
    \centering
    \includegraphics[height=3.3cm]{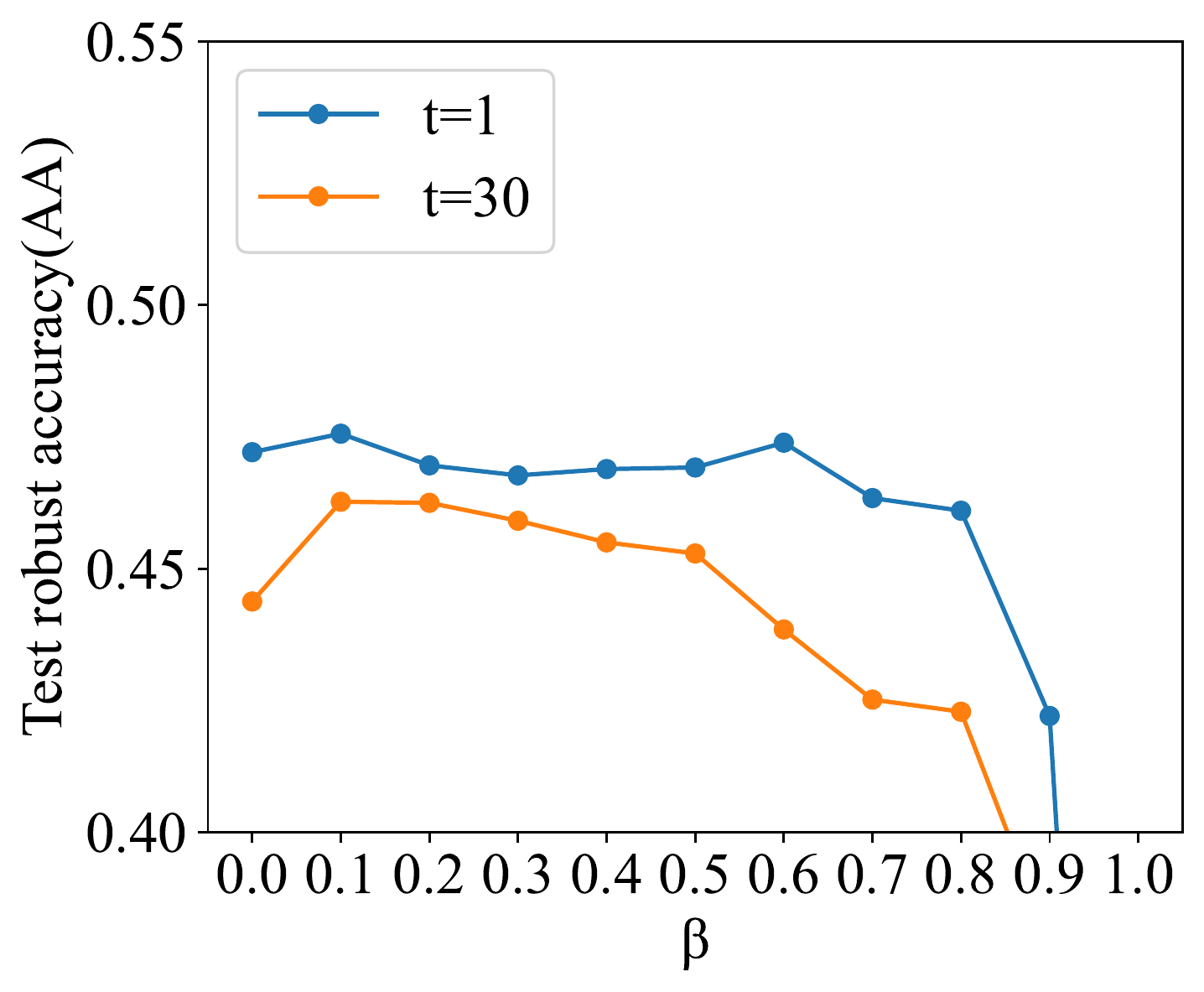}
    \subcaption{RA}
    \label{cifar10_RA}
  \end{minipage}
  \caption{Test robust accuracy on CIFAR10 against AutoAttack~(AA) in each $\beta$.}
  \label{hparam_search_cifar10}
\end{figure}

\begin{figure}[t]
   \begin{minipage}[b]{0.32\linewidth}
    \centering
    \includegraphics[height=3.3cm]{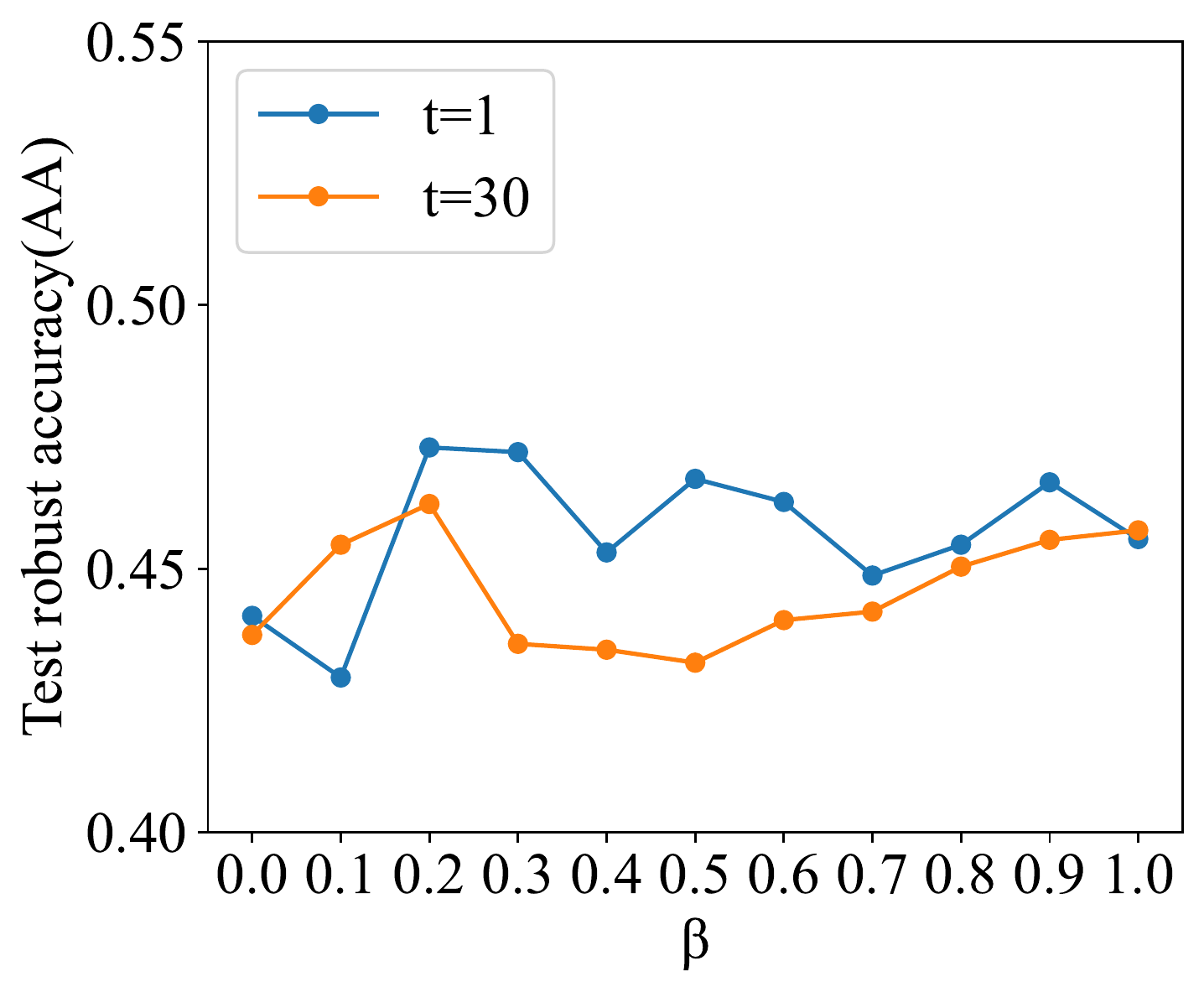}
    \subcaption{CC}
    \label{svhn_CC}
  \end{minipage}
   \begin{minipage}[b]{0.32\linewidth}
    \centering
    \includegraphics[height=3.3cm]{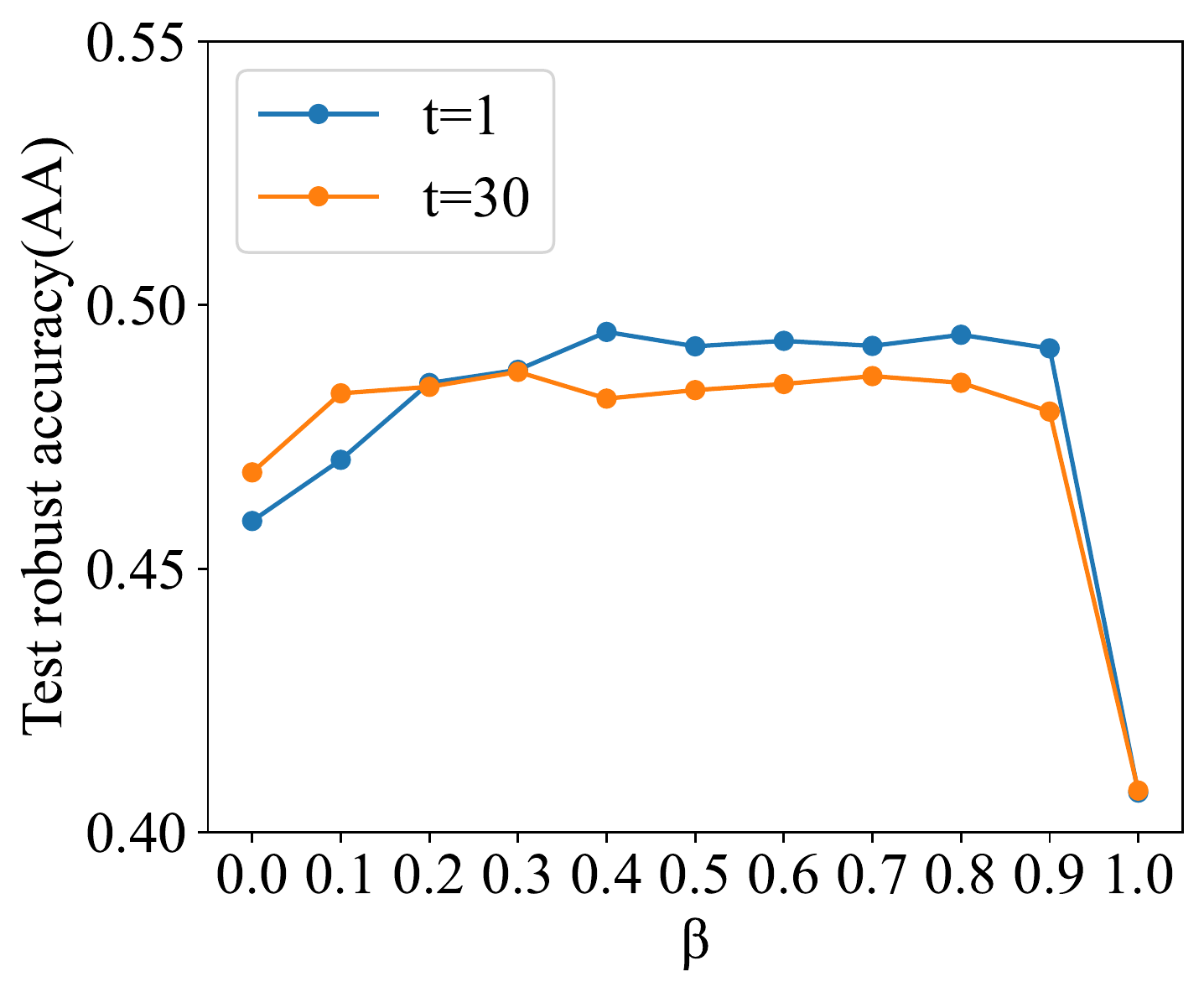}
    \subcaption{RC}
    \label{svhn_RC}
  \end{minipage}
  \begin{minipage}[b]{0.32\linewidth}
    \centering
    \includegraphics[height=3.3cm]{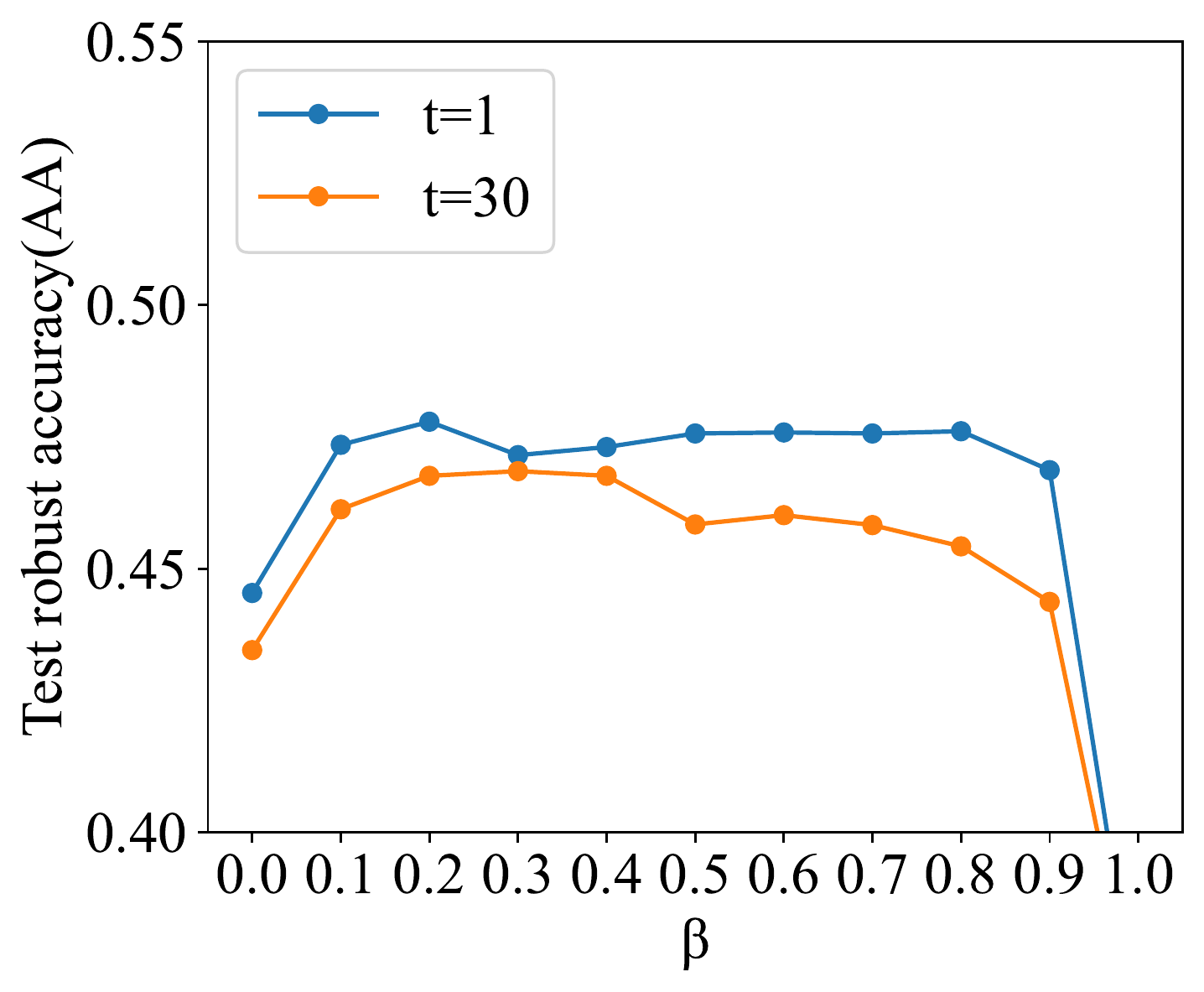}
    \subcaption{RA}
    \label{svhn_RA}
  \end{minipage}
  \caption{Test robust accuracy on SVHN against AutoAttack~(AA) in each $\beta$.}
  \label{hparam_search_svhn}
\end{figure}

\subsection{Optimization of $\beta$}
\label{sec_hyper_parameter}
In this section, we discuss the choice of hyperparameter $\beta$ for ARDIR.
For each teacher combination in the proposed method, the test robust accuracy against AutoAttack when $\beta$ is varied from $0.0$ to $1.0$.
When $\beta=0$, this is equivalent to ARD.
The results are shown in Figures~\ref{hparam_search_cifar10} and \ref{hparam_search_svhn}.
For all combinations of teacher models and data, the test robust accuracy of the student model is improved by using the internal representation. 
Since ARDIR does not use the correct label to train models when $\beta$ close $1$, its performance degradation is obvious.
Thus, we do not mention it.

First, we focus on the result of RC.
RC utilized the internal representation obtained when clean data is input into a robust model.
Since this internal representation is the informative label, RC performed better than the other combinations when $\beta=0.6$.

Second, We focus on the result of CC.
CC~($beta=0.4$) achieved better robust accuracy than without internal representation.
However, its robustness is lower than RC.
The reason is that its internal representation is not robust, because the clean model is vulnerable to the adversarial example.

Third, we focus on the result of RA.
The robust model tends to overfit the adversarial example included in the training dataset.
Therefore, the robust teacher model has a low generalization performance to the test adversarial example. 
In other words, the internal representations of the robust teacher model have a low generalization performance against the test adversarial example.
It leads to the performance degradation of the proposed method.
In fact, RA has less performance improvement than RC.

Finally, We discuss dependence on $\beta$ of performance gains when using internal representations.
Since CC and RA used poor internal representations, performance gains are marginal.
On the other hands, since RC used good internal representations, RC has greater performance gains than RA and CC when increasing $\beta$.
In addition, the performance of RC is stable with respect to changes in $\beta$, although it is best at $\beta=0.6$.

\subsection{Performance evaluation}
\label{sec_eval_perfomance}
In this section, we evaluate the performance of the proposed method ARDIR and the previous methods.
The hyperparameters for ARDIR were set as investigated in the previous section.
The hyperparameter $t$ for ARD was chosen to be the highest test robust accuracy against AutoAttack.
The performance against each attack method is shown in Tables~\ref{eval_cifar10} and \ref{eval_svhn}.
In Table~\ref{eval_cifar10}, ARD~(RA) with the robust model as the teacher model and adversarial example as the teacher data showed a very high robust accuracy for PGD.
On the other hand, ARD~(RA) has low performance against AutoAttack.
This is considered to be gradient obfuscation, and the performance to non-PGD is degraded.
On the other hand, the proposed method ARDIR~(RC) has the next highest robust accuracy for PGD after those methods, while it has higher performance for AutoAttack than the state-of-the-art method IAD.

The same trend can also be seen in Table~\ref{eval_svhn}.
ARD~(RA) and ARDIR~(RA) have high robust accuracy against PGD, but their performance against AutoAttack is lower than ARDIR~(RC).
In contrast, ARDIR~(RC) shows the highest performance in both PGD and AutoAttack.

These results show the effectiveness of the proposed method ARDIR.
As we expected ARDIR~(RC), which uses the internal representation of the robust model for clean data, has high robustness.

\begin{table}[t]
\begin{center}
\caption{Test accuracy on CIFAR10.}
  \begin{tabular}{lllll}
\toprule Defense Method & Clean &FGSM&PGD & AA\\ 
   \cmidrule(r){1-1} 
   \cmidrule(r){2-2} 
   \cmidrule(r){3-3} 
   \cmidrule(r){4-4}
   \cmidrule(r){5-5}
   SAT & 0.8094 & 0.5657&0.5229 & 0.4787 \\ 
    ARD~(CC)($t=1$) & 0.8177 & 0.5712&0.5257 & 0.4800\\ 
    ARD~(RC)($t=1$) & 0.8314 & 0.6057&0.5396 & 0.4824\\ 
    ARD~(RA)($t=1$) & 0.8246 & 0.6183&0.5834 & 0.4721 \\ 
    IAD-1 & 0.8568 & 0.6005&0.5231 & 0.4847 \\
    IAD-2~\cite{zhu2022reliable} & 0.8321 & 0.6354&0.5185 & 0.4858\\ 
    ARDIR~(CC)($\beta=0.4,t=1$) & 0.8556 & 0.6049&0.5214 & 0.4901 \\ 
    ARDIR~(RC)($\beta=0.6,t=1$) & 0.8267 & 0.6042&0.5479 & 0.5014\\ 
    ARDIR~(RA)($\beta=0.1,t=1$) & 0.8177 & 0.5877&0.5525 & 0.4756\\ 
    \bottomrule
  \end{tabular}
  \label{eval_cifar10}
\end{center}
\end{table}

\begin{table}[t]
\begin{center}
\caption{Test accuracy on SVHN.}
  \begin{tabular}{lllll}
    \toprule Defense Method & Clean &FGSM& PGD & AA\\
    \cmidrule(r){1-1} 
   \cmidrule(r){2-2} 
   \cmidrule(r){3-3} 
   \cmidrule(r){4-4} 
   \cmidrule(r){5-5}
    SAT & 0.9067 & 0.6175&0.5247 & 0.4494\\ 
    ARD~(CC)($t=1$) & 0.8977 & 0.6097&0.5150 & 0.4411 \\ 
    ARD~(RC)($t=30$) & 0.9034 & 0.6356&0.5257 & 0.4683 \\ 
    ARD~(RA)($t=1$) & 0.9113 & 0.6453&0.5600 & 0.4454 \\ 
    ARDIR~(CC)($\beta=0.2,t=1$) & 0.9299 & 0.6676&0.5450 &0.4730\\ 
    ARDIR~(RC)($\beta=0.8,t=1$) & 0.9114 & 0.6586&0.5747 & 0.4944\\ 
    ARDIR~(RA)($\beta=0.2,t=1$) & 0.9089 & 0.6462&0.5695 & 0.4779\\
    \bottomrule
  \end{tabular}
  \label{eval_svhn}
\end{center}
\end{table}
\section{Related Works}
\label{sec_relatedworks}
The various adversarial training methods were proposed~\cite{madry2017towards,wang2019convergence,zhang2019theoretically,rice2020overfitting,wu2020adversarial}.
These methods are trained using only the one-hot labels, they have been well studied.
On other hand, adversarial training using knowledge distillation utilizes the output of the teacher model as the label.
In recent research, knowledge distillation~\cite{hinton2015distilling} is gaining attention as a new approach to improve adversarial training.
The initial motivation for introducing distillation methods into adversarial training is to compress the size of robust models.
Adversarial Robust Distillation (ARD)~\cite{goldblum2020adversarially} is the first application of knowledge distillation to adversarial training.
ARD uses the output obtained by inputting clean data into a large robust teacher model as a label to train a small student model, thereby passing on the robustness of the teacher model to the student model.
Robust Soft Label Adversarial Distillation (RSLAD)~\cite{zi2021revisiting} also used the output of the teacher model as a label to simultaneously optimize the output of the student model against the adversarial example and the clean data.
In this way, RSLAD efficiently passes on the high robustness of the huge teacher model to the small student model.

One interesting phenomenon in the ARD paper is that the student model outperformed the teachar model when the same architecture was used for both the teacher and student models.
As a result, methods have been proposed to improve performance by applying knowledge distillation to adversarial training, and the proposed method is one of them.
AKD$^2$~\cite{chen2021robust} used distillation as a regularization for adversarial training to prevent overfitting and improve performance.
Introspective Adversarial Distillation (IAD)~\cite{zhu2022reliable} is the state of the art in adversarial training aimed at improving performance by knowledge distilling across the same architecture.
IAD focuses on the fact that the output of the teacher model does not work as a correct label for the adversarial example as the learning progresses.
Therefore, IAD improved the performance by using the output of the student model itself as a label in addition to the output of the teacher model.
Compared with these methods, ARDIR uses the internal representation as a more informative label.
This enables ARDIR to learn more robust student models than IAD.
\section{Conclusion}
\label{sec_conclusion}
In this paper, we investigated the performance improvement of adversarial training by knowledge distillation.
We proposed adversarial training using knowledge distillation with internal representation, called ARDIR.
Our proposed method can use more informative and generic features as labels than the conventional methods by using the internal representation of clean data input to the teacher model.
We also inspected the combination of the teacher model and data for knowledge distillation in adversarial training.
As a result, the combination of robust model and clean data was shown to be the most effective in generating a robust student model against the adversarial examples.
As we predicted, this result revealed that the internal representation obtained when clean data is input into a robust model has good information as a label.
Furthermore, experiments on multiple datasets showed that our proposed method outperforms the previous methods.

\bibliographystyle{splncs04}
\bibliography{egbib}
\end{document}